\documentclass[letterpaper, 10 pt, conference]{ieeeconf}  
\IEEEoverridecommandlockouts
\usepackage{siunitx}
\usepackage{physics}
\usepackage[T1]{fontenc}
\AtBeginDocument{\RenewCommandCopy\qty\SI}
\usepackage{amsmath,amsfonts}
\usepackage{array}
\usepackage{cite}
\usepackage[caption=false,font=scriptsize]{subfig}
\usepackage{textcomp}
\usepackage{stfloats}
\usepackage{float}
\usepackage{url}
\usepackage{verbatim}
\usepackage{graphicx}
\usepackage{amssymb}
\usepackage{pifont}
\usepackage{algorithm}
\usepackage{algpseudocode}
\usepackage[disable]{todonotes} 
\usepackage{hyperref}
\usepackage[capitalize]{cleveref}
\usepackage{comment}
\usepackage{multirow}
\usepackage{booktabs}
\usepackage{rotating}
\usepackage{makecell}
\usepackage{tabularx}
\usepackage{xcolor}
\usepackage{textcomp} 

\usepackage{blindtext}

\def\BibTeX{{\rm B\kern-.05em{\sc i\kern-.025em b}\kern-.08em
    T\kern-.1667em\lower.7ex\hbox{E}\kern-.125emX}}

\begin{document}

\title{\LARGE \bf
Accelerating Autonomy: Insights from Pro Racers in the Era of Autonomous Racing - An Expert Interview Study
}

\author{Frederik Werner, René Oberhuber, Johannes Betz 
\thanks{F. Werner, R. Oberhuber, J. Betz are with the Professorship of Autonomous Vehicle Systems, TUM School of Engineering and Design, Technical University of Munich, 85748 Garching, Germany; Munich Institute of Robotics and Machine Intelligence (MIRMI), corresponding author: frederik.werner@tum.de}
}

\onecolumn

\begin{center}
    \textcopyright \ 2024 IEEE. Personal use of this material is permitted. Permission from IEEE must be obtained for all other uses, including reprinting/republishing this material for advertising or promotional purposes, collecting new collected works for resale or redistribution to servers or lists, or reuse of any copyrighted component of this work in other works.
\end{center}

\twocolumn

\maketitle


\begin{abstract}

This research aims to investigate professional racing drivers’ expertise to develop an understanding of their
cognitive and adaptive skills to create new autonomy algorithms. An expert interview study was conducted with 11 professional race drivers, data analysts, and racing instructors from across prominent racing leagues. The interviews were conducted using an exploratory, non-standardized expert interview format guided by a set of prepared questions.
The study investigates drivers' exploration strategies to reach their vehicle limits and contrasts them with the capabilities of state-of-the-art autonomous racing software stacks. Participants were questioned about the techniques and skills they have developed to quickly approach and maneuver at the vehicle limit, ultimately minimizing lap times. The analysis of the interviews was grounded in Mayring's qualitative content analysis framework, which facilitated the organization of the data into multiple categories and subcategories.
Our findings create insights into human behavior regarding reaching a vehicle’s limit and minimizing lap times. We conclude from the findings the development of new autonomy software modules that allow for more adaptive vehicle behavior. By emphasizing the distinct nuances between manual and autonomous driving techniques, the paper
encourages further investigation into human drivers’ strategies to maximize their vehicles’ capabilities.
\end{abstract}

\section{INTRODUCTION}

The chess match between Kasparov and the computer DeepBlue served as the first notable showdown in the man versus machine competition \cite{Campbell.2002}. Since then, world champions from various disciplines have been beaten by a computer system. Most recently, in multiplayer racing in the simulated environment of Gran Turismo~7 \cite{Wurman.2022}, or under real physical conditions in drone racing \cite{Kaufmann.2023}. In the field of autonomous motorsport, it has been demonstrated that Autonomous Vehicle Software Stacks (AVSS) can operate at the level of amateur human drivers\cite{Hermansdorfer.5202020, Kegelman.2018}. Still, AVSS have not yet proven to be able to drive lap times at world champion level. Demonstrating such performance would indicate that autonomous vehicle technology is sufficiently mature to manage emergency situations in public driving environments as well.
The Indy Autonomous Challenge (IAC) \cite{Raji.2023, Betz.2023} recently showcased an autonomous racing performance on the Formula~1 track in Monza, revealing that AVSS have not yet surpassed their human counterparts. The fastest recorded lap time for the autonomous racecar was 2:05,87~\cite{Indy2023}, with no laps completed by a human in an identical vehicle for comparison. For context, in the human-driven Formula 4 championship 2023, a Tatuus F4-T421 achieved a lap time of 1:52.86. Notably, the Tatuus features a lower power-to-weight ratio compared to the Dallara IL15 used in the IAC.

\begin{figure}
\centering
\includegraphics[width=1\columnwidth]{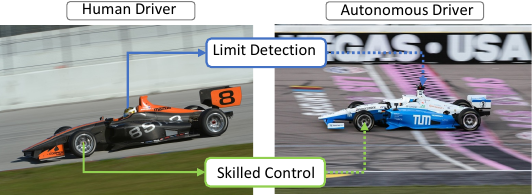}
\caption{A human driver (left) on the Dallara DW12 chassis in comparison to an autonomous driver (right) on the Dallara IL15 chassis. We can learn from human expertise to understand the necessary tasks and cognitive skills to detect the limit and apply skilled control.}
\label{fig:Introduction}
\vspace{-1em}
\end{figure}

This example indicates that current AVSS can navigate new environments with sparse data availability. However, AVSS are still clearly outperformed by human drivers.A professional race driver is capable to swiftly improve their lap times within a few laps and get close to the vehicle’s performance potential. Human intuition, especially in complex tasks like race driving, is built upon years of experience, sensory integration, and adaptive learning. Humans possess the ability to understand subtle cues in the environment, anticipate the actions of other drivers and the car, and make split-second decisions at the limits of handling based on a combination of current observations and past experiences. We therefore see the need to elicitate the knowledge of professional racing drivers to uncover the cognitive and adaptive skills crucial for behaviour adaption due to unknown and evolving vehicle limitations as well as skilled vehicle control \cref{fig:Introduction}.

In summary, this work has three main contributions:
\begin{enumerate}
    \item We present the results of an expert interview study amongst 11 professional race drivers, data analysts, and racing instructors across prominent racing leagues. 
    \item We display the human drivers' exploration strategies to reach their vehicle limits and contrasts them with the capabilities of state-of-the-art autonomous racing software stacks.
    \item We distill insights from the comparison and generate concepts to enhance the performance of modular AVSS in autonomous motorsport by drawing inspiration from human drivers.
\end{enumerate}

\section{RELATED WORK}
\subsection{Various racing software architectures}
The publications related to racing AVSS can be categorized into two principal domains. On the one hand, end-to-end machine learning (ML) approaches exist that conceptualize the entire driving function as an integrated ML problem, on the other hand there are modular hierarchical approaches.

End-to-end approaches use Imitation Learning (IML) or Reinforcement Learning (RL) \cite{Cai.2021}. RL approaches offer advantages such as the semi-automated acquisition of the necessary skills through self-learning and the potential attainment of superhuman-like capabilities regarding precision and consistency \cite{Fuchs.2021, Wurman.2022}. However, those approaches use a large observation space with direct feature extraction non-transferable into reality, as criticized by Samak et al.~\cite{Samak.10112021}. The most common limitation of IML approaches is the poor generalization to out of distribution data and the inherent restriction by the quality of the labeled dataset~\cite{Samak.2021}. In contrast, RL approaches suffer from the substantial data required, limited safeguarding possibilities, and a limited capacity to generalize to novel environments. In practical applications, such approaches find prevalence in simulation due to their comparatively cost-effective data generation possibilities or in small-scale vehicles enabled by their lower operational and accident costs \cite{Cai.2020, Fuchs.2021, Wurman.2022, Samak.10112021}.

\begin{figure}[h]
\centering
\includegraphics[width=0.48\textwidth]{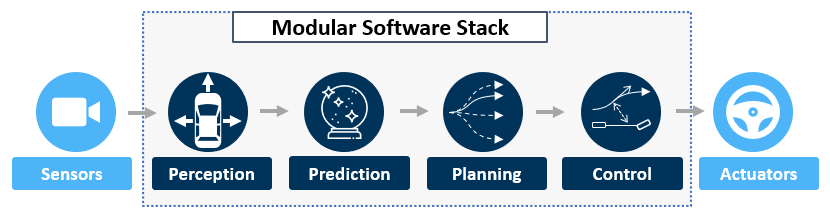}
\caption{The basic autonomous driving software architecture of a modular AVSS designed for autonomous racing \cite{Betz.2022}}
\vspace{-1em}
\label{fig:software stack}
\end{figure}

Modular, hierarchical approaches follow a sense-plan-act structure often labeled as perception, planning, and control \cite{Betz.2022} \cref{fig:software stack}. 
While individual modules may incorporate ML methods, their application is typically confined to the scope of the module. Feedback in the opposing direction to the sense-plan-act data flow is rare. This approach is commonly favored in real-world full-size vehicles due to the reasonable possibilities for implementing safety concepts and the lower data demand, stemming from the option to substitute data-driven modules with more classical approaches~\cite{Raji.2023, Betz.2023}. Common difficulties include high computational demand, the risk of fault propagation, and performance degradation due to model mismatch \cite{Samak.10112021, Betz.2023, Raji.2023}. In a racing simulation, Hao et al. \cite{Hao.11172022} demonstrated that human drivers can be outperformed even with modular approaches. However, this was achieved by leveraging the simulation's stable physics with precise model identification and manual iterative tuning. 
Modular approaches often face challenges leveraging the self-learning capabilities inherent to end-to-end systems, which seems to be the key to reaching superhuman performance~\cite{Fuchs.2021,Wurman.2022}. Therefore, it is crucial to investigate the 'speed secrets' of human drivers, who continue to set the benchmark in real-world racing in order to set the right development goals for modular AVSS. 

\subsection{Research on human driver behavior for autonomous software design}
Looking into related research in this field, Kegelman~\cite{Kegelman.2018} explored how skilled human drivers operate vehicles at their limits and compared these findings with modular AVSS. He included a detailed analysis of racing trajectories, driver behavior, and vehicle stability, concluding that human drivers excel in the continued maximization of the resulting accelerations, even in unstable vehicle conditions. Hermansdorfer et al. \cite{Hermansdorfer.5202020}  compared a professional human driver and a modular AVSS on the Roborace autonomous racing vehicle showing distinct differences in the driven path, acceleration archived, and smoothness of vehicle inputs. The authors pointed out that the most significant lap time improvements could be achieved with an online adaptable approach to the acceleration constraints, track boundary safety margins, and a robust control concept able to stabilize the vehicle at the limit. Remonda et al.~\cite{Remonda.2021} investigated human drivers and RL agents in a simulator, demonstrating that the two groups achieved similar lap times. However, they did so with different characteristics and each group exhibited unique features crucial for their lap time performance. The authors suggested that incorporating certain human driving behaviors could be beneficial in training RL agents. Doubek et al. \cite{Doubek.2021} did an interview study that delved into the qualities that constitute good driving on public roads and race tracks. Key findings indicated good driving on racetracks involves adaptability of the used race line to the vehicle and non-aggressive vehicle control resulting from drivers' confidence and experience.

\section{INTERVIEW DESIGN AND EVALUATION}
We conducted expert interviews with professionals from various sectors of the racing industry. The primary objective of the interviews is to address the question: "How does a professional human race car driver operate a vehicle at its limits?" To answer these questions, we are using an exploratory approach to gather in-depth insights into the cognitive and adaptive skills of a human race driver. Experts are expected to provide detailed insights on how vehicles behave when approaching their limits, sharing personal experiences and techniques.

The interviews are designed to last 60 minutes, employing a standardized interview guide. The interviews were conducted in person or via video call and were audio-recorded. The experts gained their relevant experience in prominent racing series such as Formula 1, Formula 3, DTM, and ADAC/GT Masters and were selected through personal recommendation. The interview guide, demographic details, selected quotations, and comprehensive paraphrasing in the original interview language (German) can be accessed via the link provided in the references~\cite{FrederikWerner.2024}. The interview guide contains 13 main questions (Q1-Q13), complemented by follow-up questions as needed. The interview guide commences with an inquiry into the individual definition of the limit (Q1) and subsequently progresses through topics such as driving behavior (Q2) and vehicle feedback at the limit (Q3) to the core question of reaching and recognizing the limit of the vehicle (Q4-Q6). Furthermore, topics such as racing lines (Q7), maximizing vehicle potential (Q8, Q12), data analysis (Q9-Q10), and rain and wet conditions (Q11) are discussed. (Q13) is used to statistically assess the relevance of vehicle parameters for exploring and analyzing the vehicle limit, distinguishing it from other questions that are qualitatively evaluated and permit freely formulated responses. The interview guide is repeatedly refined between the interviews to enhance the clarity of the questions and incorporate insights. Demographic data is gathered using a separate pre-interview questionnaire. A post-interview protocol is conducted to document the overall conversational atmosphere.

\subsection{Evaluation}
\label{subsec:Evaluation}
\begin{figure}
\centering
\includegraphics[width=0.45\textwidth]{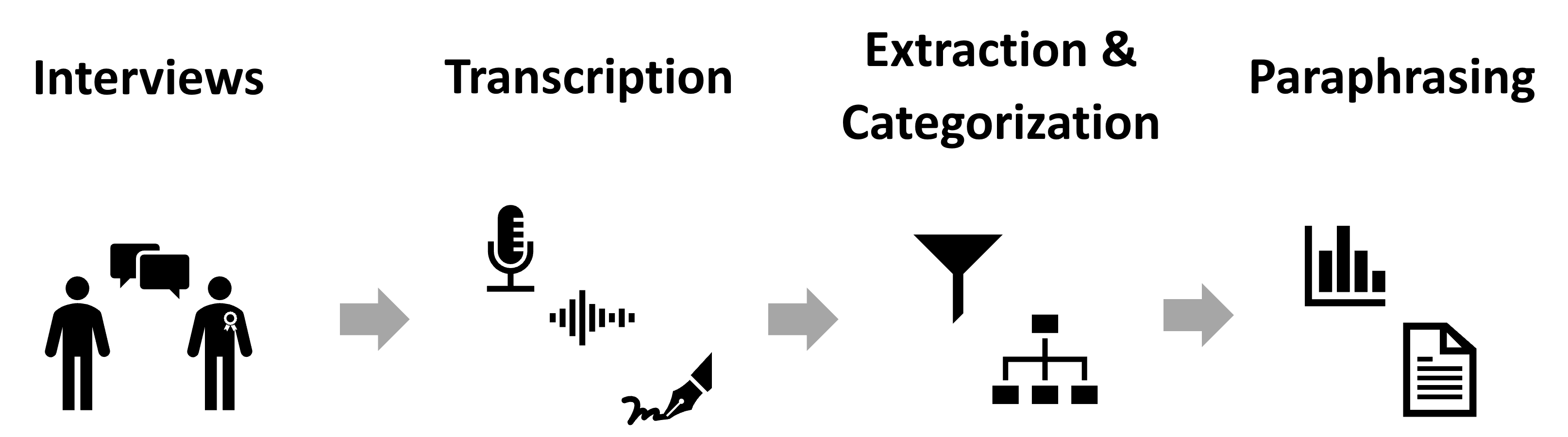}
\caption{The evaluation framework employed during this study}
\label{fig:process}
\end{figure}

The expert interviews are evaluated following the \textit{qualitative content analysis} method proposed by Mayring \cite{Mayring.2000}. An adaptation presented by Kaiser \cite{Kaiser.2014} encapsulates the essential steps of this analytical methodology for application in theory-generating and exploratory expert interviews, serving as the foundation for the analysis. The steps are delineated below and visualized in Figure \ref{fig:process}.
\begin{enumerate}
\item \textbf{Interviews \& transcription:}
All interviews are acoustically recorded using two independent recording devices. The comprehensive transcription facilitates a transparent execution of the qualitative content analysis. Anonymizing the interview involves excluding names, projects, and companies and assigning it an anonymous code in the form of \#0XX.
\item \textbf{Extraction \& categorization:}
The analysis commences with identifying content-relevant quotations in respect to the research question, followed by designing a category system based on the identified quotations to group thematically similar statements. The authors independently created individual category systems, which were subsequently cross-examined and harmonized, leading to the establishment of a validated category system.
\item \textbf{Paraphrasing:}
In the final step of qualitative content analysis, all quotations of a subcategory are paraphrased into a general statement. This decouples the results from the quotations, enabling them to be treated as independent theories and facts. The paraphrasing at the subcategory level provides more points of connection for further analysis and interpretations that can be revisited in subsequent investigations. Following paraphrasing, core statements of each subcategory are formulated, aiming to concisely represent the thematic area and illustrate the content of the paraphrase.
\end{enumerate}

\section{RESULTS}
As a result, we conducted eleven expert interviews totaling 14.5 hours of audio recordings. Among the eleven experts, seven are of the class race drivers, four are instructors, and five are vehicle dynamics experts. Overlapping designations are possible, as some individuals, for example, former race drivers, now serve as instructors. ~\cref{table:demographics} displays the summary of participants' demographic data.

\begin{table}[h]
\centering
\caption{Demographic data of this study's participants}
\begin{tabular}{|c|c|c|}
\hline
\textbf{Variable} & \textbf{Unit} & \textbf{Value}\\ \hline\hline
\textbf{Participants} & count & 11 \\ 
\textbf{Gender} & count & 10 (m) / 1(f) \\
\textbf{Age} & years & 32-59 / \O42.7 \\ 
\textbf{Relevant Experience} & years & 2-48 / \O20.4 \\
\hline
\end{tabular}
\label{table:demographics}
\end{table}

 In the analysis, due to the study's exploratory nature, no further distinction was made among the expert classes. The mention of the expert classes indicates that the topic is examined from various perspectives. The interviews resulted in 306 pages of transcriptions, yielding 370 selected quotations. The quotations were grouped into eight main categories, further divided into 30 subcategories following the framework described in \cref{subsec:Evaluation}. The subsequent step involved paraphrasing and condensing the information. 
 
\begin{figure}[H]
\centering
\includegraphics[width=0.45\textwidth]{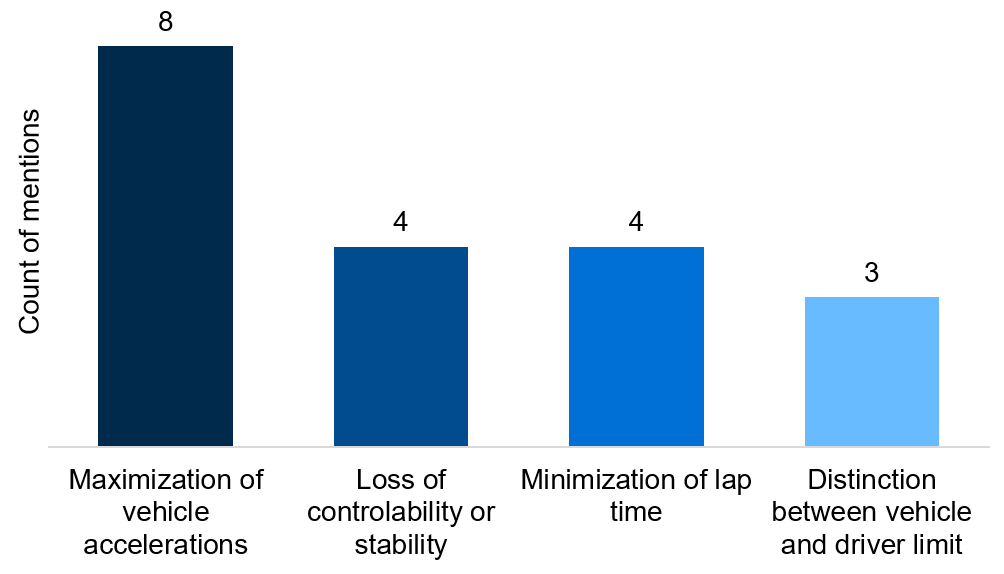}
\caption{Evaluation of (Q1): Most common mentions of the experts when asked about their personal definition of the \textit{vehicle limit}.}
\label{fig:Limit}
\end{figure}

The initial question aimed to establish a standard definition for the term \textit{vehicle limit} which is essential for the following discussions. Experts first articulated the \textit{vehicle limit} in their own terms, and subsequently, the interviewer provided a textbook definition to enhance clarity. The textbook definition for \textit{vehicle limit} was given as: \textit{The limit describes the area of the combination of longitudinal and lateral acceleration just before the friction limit between the road and tires, and is thus, for example, a limiting factor in terms of the maximum lateral acceleration during cornering} \cite{Kegelman.2018}.
Essentially, the experts agree with this, but several additional descriptions are frequently made, as seen in \cref{fig:Limit}.

\begin{figure*}
\centering
\includegraphics[width=0.9\textwidth]{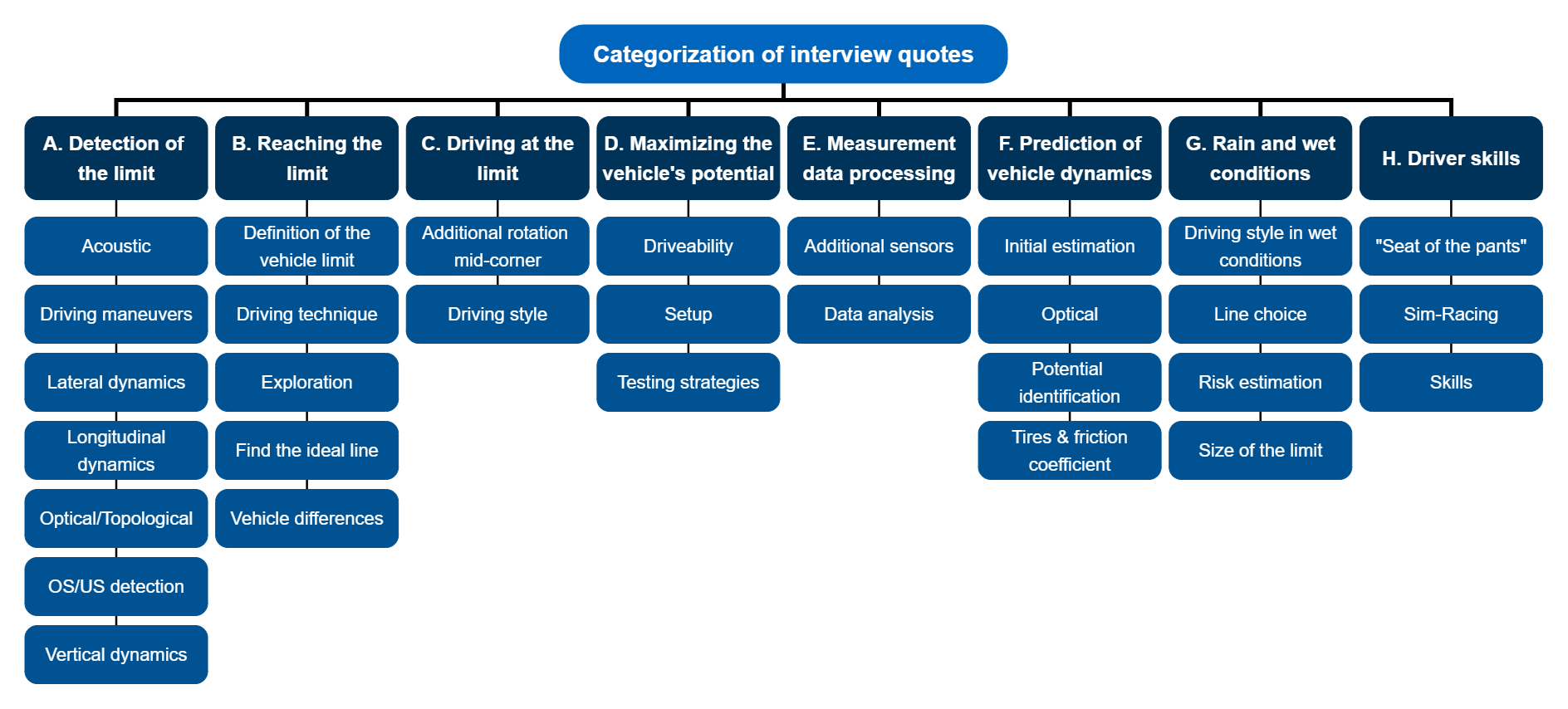}
\caption{The derived eight categories and 30 subcategories, following the application of the analysis framework. Quotations from the interview, along with their categorization and corresponding paraphrasing, are detailed in \cite{FrederikWerner.2024}. A condensed version of the paraphrasing is given in \cref{sec:analysis}}
\label{fig:categories}
\end{figure*}

The widely accepted rationale for the vehicle limit indicates that the car is operated near the boundary of Kamm's Circle, particularly when the vehicle's tires are close to being fully utilized with lateral and longitudinal forces. Here, the vehicle's behavior diverges from that within the inner area of the circle, typically indicated by a reduction in controllability or stability ~\cite{Milliken.1995}. Moreover, reaching the vehicle limit is often signified by the minimization of lap time, another commonly cited indicator. An intriguing point raised was the differentiation between the \textit{driver limit} and the \textit{vehicle limit}. The limit for a driver is defined by the experts as the point where the driver can no longer maintain control of the vehicle. This contrasts with the vehicle limit, which may vary, as another driver might successfully control the vehicle under similar conditions.

\section{ANALYSIS \& DISCUSSION}
\label{sec:analysis}

We now summarize the findings of the interview questions, consolidated into eight primary categories (A-H) shown in~\cref{fig:categories}. For the paraphrasing according to the subcategory structure please refer to \cite{FrederikWerner.2024}. After each category, we discuss the relevance to the development of advanced AVSS for autonomous racing applications.

\subsection{Detection of the limit}
The initial category of the interview findings addresses all vehicle feedback or driving strategies that render the dynamic limits perceptible and visible to the driver and dives into the question of "How do you recognize that you are operating close to the limit?".

Our study reveals that drivers employ different methods to detect operation at the limit. On certain tires, acoustic cues like tire squeal can indicate nearing the tire's maximum potential. Additionally, sudden pitch increases reveal the maximum friction $\mu$ of the driven axle has been surpassed. Drivers also use identification maneuvers like quick steering movements to gauge the vehicle's reaction at the limit. Brake pedal feedback and visual feedback from electronic aids like Traction Control (TC) or Anti-lock Braking System (ABS) help detect the dynamic limit at corner entry and exit. Understeer and oversteer detection play a vital role in detecting the limit. Drivers recognize understeer by the loss of steering torque on the steering wheel. Oversteer is perceived through over-rotation and the increase in yaw acceleration. Another common mention is \textit{anomaly detection}: If the vehicle's reaction significantly differs from the anticipated behavior, it indicates that the limit has been reached. 

\textbf{Discussion:} In this category, several connections to the area of friction coefficient $\mu$ estimation algorithms can be drawn. There are a variety of approaches, including those based on acoustics, wheel alignment torque, tire slip, or vibrations \cite{Acosta.2017, Khaleghian.2017}. However, drivers use these techniques primarily in areas very close to the limit and to assess on which level relative to the limit they were operating. A correspondingly strong behavioral adjustment is triggered depending on the perceived relative level. Human drivers' strength lies in combining many different detection techniques to enable robust limit detection. Research on integrating different limit detection mechanisms into a fusion algorithm for robust limit detection is advised as a topic for future research.

\subsection{Reaching the limit}
The path to reach the vehicle limit can take various forms. In this category, human drivers describe their strategies for exploration, revealing distinctions between various types of vehicles. Driving technique and the individual ideal racing line are crucial for consistent operation at the limit.

Our study reveals that exploring a vehicle's limit involves drivers comparing its usual stable behavior with its current behavior and making increasingly stronger inputs to assess its characteristics. If the vehicle responds correctly without understeering or oversteering, the limit has not been reached, leading to adjustments in braking points and cornering speed. Drivers test braking points and track limits in initial laps to determine, which curbs can be used for shortening the track or added rotation. The most challenging aspect of cornering is achieving optimal apex speed, which depends on finding the right braking point, managing brake pressure, and adjusting the steering angle to avoid understeering. At the limit, each vehicle reacts differently to driver inputs and can change its behavior throughout a session. As a result, race drivers adjust their driving techniques and race line style to match the vehicle characteristics. Commonly, the main focus is on maximizing corner exit speed to reduce lap time. These adjustments are validated in real-time using predictive lap time models or delta time indicators.

\textbf{Discussion:} The available tire grip is subject to various external factors and changes continuously. In planning and control, reliance is still placed on a measure of the available tire grip, commonly denoted as the coeffficient of friction $\mu$. Based on an estimation of tire limitations, the algorithms of planning and control can be constrained to output dynamically feasible trajectories and control commands.~\cite{Betz.2023}. To safely operate at the limit, a precise estimation is required. Tire-road friction estimation approaches usually struggle with high uncertainties, improving when increasingly excited \cite{Acosta.2017, Khaleghian.2017}. Even with a theroretically perfect estimate, unmodeled effects within the planning or control module may prevent consistently reaching the limit, leading to ever increasing model identification requirements and computationally more expensive algorithms. Consequently, an intriguing research question emerges to circumvent the escalating requirements for model fidelity and parameter identification: Should the approach be altered to emulate human drivers who progressively explore and observe vehicle behavior to approach the limit? 

Another interesting observation is that race drivers intentionally test and exploit the use of curbs or cutting of corners to their advantage in either reducing curvature of the raceline, raceline length, or triggger desired vehicle reactions. For AVSS, the lateral deviation is a commonly encountered state variable that describes the control error relative to a target trajectory in the Frenet coordinate system. However, it is inherently an open-state variable, not an optimization variable: achieving precise apex targeting and fully exploiting the drivable area typically involves an iterative manual process between racing line optimization and control tuning. Moreover, track boundaries are not always clearly defined. While some have distinct lines, there are also curbs of varying heights, where the advantage in lap time becomes apparent only through traversing them and observing the resulting vehicle's response. With unavoidable path tracking deviations, AVSS struggle with the maximum exploitation of the drivable area as mentioned in \cite{Hermansdorfer.5202020}. A targeted, self-optimizing approach for maximizing the utilization of track boundaries is an area for further research. 

\subsection{Driving at the limit}
Handling the ensuing vehicle responses becomes crucial upon reaching the vehicle dynamic limit. The next section provides a summary of the methodologies employed by drivers for this purpose.

The results of our interviews indicate that racing drivers often identify the near-perfect braking point on initial attempts using visual cues and experience, aiming for maximum lateral acceleration at the apex. This is achieved through visual markers like signs or track features, which help maintain consistent driving lines. Trail braking up to the apex is deemed most efficient during the braking phase. Drivers strategically shift braking points, exploring grip through line variations in the braking zone. At the apex, the focus is on avoiding an unloaded state by maintaining tension in the vehicle through slight braking while accelerating or starting to turn early, with different styles like late and sharp turns (v-style) or constant circular motion (double apex). 
Additional rotation techniques are vital for maintaining performance close to the limit. Load changes, resulting in axle load transfer, enable further yaw rotation boosts mid-corner. This is achieved without additional steering input, often through brief brake taps or momentarily releasing the accelerator, triggering agile vehicle turning, and is used especially to overcome understeering tendencies at the limit. Track features such as curbs also contribute to inducing this additional rotation. Strong throttle inputs are used in understeering vehicles to induce oversteer.
At the corner exit, the steering is opened proportionally to the accelerator pedal input, and the vehicle's position is considered depending on the subsequent course of the track. Once a racing driver commits to accelerating, there is no going back. Instead, they adjust through other variables such as steering angle and line choice, accepting deviations from the ideal line if the upcoming section is suited. 

\textbf{Discussion:} If a too-fast trajectory is requested by the planning module, it can lead to the saturation of the tires, manifesting as limit oversteer (LOS) if the rear axle saturates first or limit understeer (LUS) when the front axle saturates. In such cases, the front axle loses its ability to manage lateral control effectively.
A problem with current control approaches is that they either use only steering for lateral guidance \cite{Wischnewski.2023b} or require precise estimation of tire friction values and parameterization to properly react to LUS \cite{Raji.2023b}. Laurense et al.~\cite{Laurense.2017} have shown that LUS can be effectively mitigated with a tire slip-based control. Thus, vehicle control at the limit is an integrated problem with an additional control objective: vehicle stability. This leads to the following question for future research: How can LOS/LUS be effectively managed while being robust against high uncertainties in friction estimation? How can a situation-aware prioritization between vehicle stability, track limits, path-tracking, and speed-tracking control objectives occur in such situations to prevent an underactuated control problem?

\subsection{Maximizing the vehicle's potential}

This section discusses how race drivers, together with their engineers, maximize the vehicle potential through setup and testing strategies.

Our study reveals that drivers' confidence in their vehicle is essential for maneuvering at the limit, with trust built on receiving unfiltered feedback from the steering, seat, and pedals. 
The vehicle setup forms the basis and should enable the vehicle dynamic limit to be reached plausibly and unambiguously. Setup loops with driver feedback are used to maximize the driver-vehicle potential and enable the driver to reach the vehicle limit consistently. This setup is customized based on the driver's preferences and feedback with continuous adjustments to address discrepancies. Changes like brake balance are often made by the driver during the race, adapting to each turn.
Testing strategies like A-B-A tests are used to evaluate the impact of new parts, setup changes, or different driving styles, helping to distinguish between driver improvements and track evolution. These tests involve running baseline laps, making adjustments or trying different driving styles, and then returning to the baseline to gauge the impact of changes.

\subsection{Measurement data processing}
The section on data processing addresses information on the utilization and analysis of telemetry data.

Our study findings indicate that data analysis is essential in motorsport for performance optimization, particularly during vehicle setup. Marker buttons on the steering wheel help capture key moments to be investigated. The analysis typically focuses on key data like steering angle, brake pressure, throttle position, speed, and lap time, which reveal insights into vehicle behavior and driver input. Steering angle data helps understand tendencies towards oversteer or understeer. Brake patterns highlight the importance of smooth transitions between braking and acceleration. Comparing data across different runs and drivers can further provide insights into vehicle and lap time potential. 

\begin{figure*}
\centering
\includegraphics[width=0.95\textwidth]{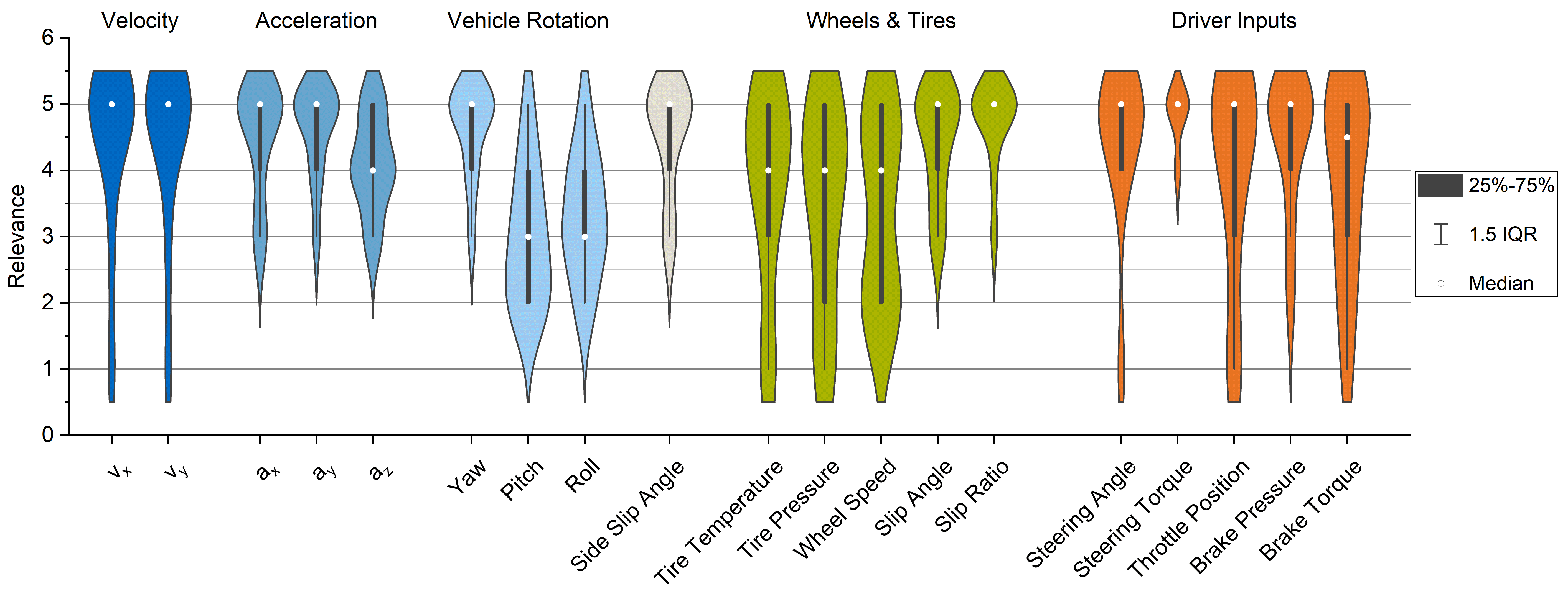}
\caption{Results of the question \textit{"Rate various vehicle signals in terms of relevance for exploring the vehicle limit"}.}
\label{fig:violin}
\vspace{-1em}
\end{figure*}

In the final question of the interview, the experts rank different measurement signals on a scale from one (not relevant) to five (absolutely necessary) regarding their relevance in the exploration of the limit. The results are shown in Figure \ref{fig:violin}. The highest relevance was attributed to vehicle speed, driver inputs as well as lateral and longitudinal accelerations, and yaw rate. Longitudinal and lateral slip signals received the highest significance as well. When asked if there were any other relevant signals the participants added downforce estimation, ride height, suspension travel, and wheel torques in vehicles featuring torque vectoring systems.

\subsection{Prediction of vehicle dynamics potential} 
In this section, drivers explain how, based on visual or haptic vehicle feedback and their experience, they assess the vehicle performance potential in the initial laps, identify additional lap time potential, and manage tire wear.

The analysis of the interview questions showed that the initial estimate is primarily based on the drivers experience and the current environmental and track conditions. A few intentional wheel load changes and the vehicle reaction during driving through the first turns are usually enough for professional drivers to find and assess the vehicle limit and vehicle characteristics. In dry conditions, it is usually faster to drive on the areas of the track that are covered with rubber abrasion and, therefore, offer more grip. Tire selection is crucial in motorsport for track performance and lap times. Tires vary in characteristics like profile, size, and compound, affecting responses to load changes and tendencies like understeer. Temperature significantly influences tire force transmission; maintaining the optimal temperature is key, as exceeding it can lead to increased wear and reduced tire potential. Slick tires, commonly used in motorsport, offer maximum grip due to their large contact area but have a small margin for error with little warning before losing traction. Increased tire pressure and temperature can cause a doughy or spongy driving feel, noticeable in the delay between steering input and vehicle response. 
Skilled drivers can judge tire wear individually for each tire. For instance, understeer in a left turn might indicate wear on the right front tire, and vice versa. Rear tire wear is usually assessed based on oversteer at the end of turns.
Estimating tire grip is considered one of the hardest aspects of vehicle dynamics modeling, as accurate estimates are only possible when the traction limit is reached. Rubber pickup, especially on slicks due to high temperatures, reduces grip and affects surface temperature, leading to grip loss. This pickup can be detected by vibrations at high speeds and is typically managed through specific driving techniques. 

\textbf{Discussion:} The quality of the initial guess about the grip condition offered is important to safe driving in the first lap where there is no reference data from previous laps available to the AVSS. Even if there was a previous outing on the same track the external conditions might change over time. External factors such as temperature, ground material, fluids or particles on the ground, and knowledge about vehicles and tires can give hints to achieve a first guess about the achievable acceleration potential of the car. Khalegian et al.~\cite{Khaleghian.2017} review different experimental friction estimation algorithms where sensors like cameras or microphones are used to correlate the sensor input to measured friction, relevant for the initial lap estimate. They conclude that the accuracy of the methods is reduced when the testing condition deviates from the condition under which the methods were trained. Future research could leverage these approaches to identify areas with varying grip levels, such as rubber marks, tire debris, and dirt. This information could be instrumental in enhancing race line planning strategies.

\subsection{Rain and wet conditions}
This section focuses on the specific differences in wet driving conditions. Not only the vehicle's limit changes, but also the applied driving style, risk assessment, and the racing line.

The experts in our interview study explained that in wet conditions, motorsport drivers start with a cautious lap to assess the reduced grip. Initially, they reduce their dynamic input by 30-50\% compared to dry conditions. Their driving style changes, becoming more aggressive and harsh and simultaneously more controlled and anticipatory. Drivers intentionally induce understeer at corner entries to maintain rear-end stability. After the initial lap, they quickly adapt and reexplore available grip, similar to dry conditions.
The limit in wet conditions is smaller, requiring drivers to react more frequently and quickly to maintain control. Reduction of lateral grip is felt stronger than longitudinal.
Drivers continuously adjust their line each lap, responding to the changing track conditions and avoiding areas with rubber deposits, which offer little traction in the wet. 
Overall, driving at the limit in wet conditions presents a heightened risk, demanding careful assessment based on the specific situation, such as session type and overall strategy goals. 

\textbf{Discussion:} Driving in the rain highlights the adaptive processes of human drivers when vehicle behavior rapidly changes due to environmental conditions and may deviate from the driver's anticipated behavior. This suggests that for AVSS, effectively managing variable grip conditions is essential, particularly in planning and control algorithms. AVSS must consistently monitor grip and adapt its planning accordingly. Wischnewski et al. \cite{Wischnewski.112019} showcased a system capable of autonomously determining vehicle limits by employing Gaussian Processes to minimize lap time by altering the velocity profile in the online planner and monitoring differences in slip angles and tire slip. The system's operation is confined by a single global scaling factor leading to a lap time limited by the one section on the track offering the lowest grip. Kapania et al. \cite{Kapania.2020} proposed a method that improves responsiveness to local variations by segmenting the track into sectors. However, this approach still lacks satisfactory resolution, and there are no published algorithms that can identify and selectively incorporate small areas with different grip availability, such as patches with rubber marks.

\subsection{Driver skills}
This subsection consolidates additional skills of race drivers that do not fit into the previously discussed main categories but are nonetheless noteworthy.

The analysis of our study revealed, that an essential skill is the "seat of the pants" feeling, where drivers use their bodies as a tool to perceive vehicle feedback. This includes sensing acceleration and rotation in all three spatial axes, haptic feedback from the steering and throttle pedals, visual cues related to yaw rate, and even auditory signals like tire squeal or curb noises. This sensory information, combined with experience, aids in taking appropriate countermeasures to vehicle reactions. Interestingly, when asked about a possible digital counterpart, many participants suggested a 6D sensor or a ML model.
Sim racing and simulator work forms an essential part of race preparation. They allow drivers to familiarize themselves with track layouts and help teams anticipate vehicle responses to setup changes. However, professional sim racers rely more on visual impressions and often lack the "seat of the pants" feeling critical in real racing, which is best developed through actual driving experience.
Additionally, experienced drivers can accurately estimate potential braking and turning points based on the visual appearance of the road surface and turn layout, along with their knowledge of the car's acceleration behavior. In summary, the abilities of racing drivers are often described as intuition based on experience. Subconsciously and almost instinctively, drivers, for example, issue steering commands after a small bump in anticipation of the vehicle skidding.

\textbf{Discussion:} Many decisions made by race drivers rely on experience and the consideration of subtle cues without a rational explanation or logical chain of reasoning. Ultimately, achieving superior performance to humans may require finding a way to emulate human intuition, as suggested by the "seat-of-the-pants" sensation experienced by drivers, or their skill in utilizing nuanced cues and multisensory input in an integrated way to guide their behavior and control of the vehicle. With end-to-end ML approaches showing early promise for this, the next step could involve incorporating these intuitive elements into hierarchical AVSS. Enabling a dynamic interplay of information and experience among system modules could create a more adaptive, experience-informed response mechanism. This complex task, if achieved, could be transformative for autonomous racing. 

\section{CONCLUSION \& OUTLOOK}

This study is, to the best of our knowledge, the first to interview professional race drivers to explore how one can learn from the human expertise to understand the necessary tasks and cognitive skills to detect the vehicle limit and apply skilled control in race cars. We conducted 11 expert interviews, categorized and analyzed the outcomes critically, and derived learnings for future autonomous vehicle development. The study's findings underscore the potential of harnessing expert insights to refine Autonomous Vehicle Software Stacks (AVSS), particularly emphasizing the need to encapsulate human intuition and adaptability within these systems. Such advancements are crucial for navigating the myriad of challenges presented by varied environmental conditions.
In the discussion section, several critical future research questions emerged, developed from contrasting current Autonomous Vehicle Software Stacks (AVSS) with insights from this study. These research questions are particularly relevant for systems with limited data availability, indicating a path to enhance overall performance and learning speed in unfamiliar or changing environments. This work bridges the gap between current technology and expert driver knowledge, paving the way for advancements in autonomous racing.

\addtolength{\textheight}{-12cm}   





\bibliographystyle{IEEEtran}
\bibliography{Bibliography}

\begin{thebibliography}{10}
\providecommand{\url}[1]{#1}
\csname url@samestyle\endcsname
\providecommand{\newblock}{\relax}
\providecommand{\bibinfo}[2]{#2}
\providecommand{\BIBentrySTDinterwordspacing}{\spaceskip=0pt\relax}
\providecommand{\BIBentryALTinterwordstretchfactor}{4}
\providecommand{\BIBentryALTinterwordspacing}{\spaceskip=\fontdimen2\font plus
\BIBentryALTinterwordstretchfactor\fontdimen3\font minus
  \fontdimen4\font\relax}
\providecommand{\BIBforeignlanguage}[2]{{%
\expandafter\ifx\csname l@#1\endcsname\relax
\typeout{** WARNING: IEEEtran.bst: No hyphenation pattern has been}%
\typeout{** loaded for the language `#1'. Using the pattern for}%
\typeout{** the default language instead.}%
\else
\language=\csname l@#1\endcsname
\fi
#2}}
\providecommand{\BIBdecl}{\relax}
\BIBdecl

\bibitem{Campbell.2002}
M.~Campbell, A.~Hoane, and F.-h. Hsu, ``Deep blue,'' \emph{Artificial
  Intelligence}, vol. 134, no. 1-2, pp. 57--83, 2002.

\bibitem{Wurman.2022}
P.~R. Wurman, S.~Barrett, K.~Kawamoto, J.~MacGlashan, K.~Subramanian, T.~J.
  Walsh, R.~Capobianco, A.~Devlic, F.~Eckert, F.~Fuchs, L.~Gilpin,
  P.~Khandelwal, V.~Kompella, H.~Lin, P.~MacAlpine, D.~Oller, T.~Seno,
  C.~Sherstan, M.~D. Thomure, H.~Aghabozorgi, L.~Barrett, R.~Douglas,
  D.~Whitehead, P.~D{\"u}rr, P.~Stone, M.~Spranger, and H.~Kitano, ``Outracing
  champion gran turismo drivers with deep reinforcement learning,''
  \emph{Nature}, vol. 602, no. 7896, pp. 223--228, 2022.

\bibitem{Kaufmann.2023}
E.~Kaufmann, L.~Bauersfeld, A.~Loquercio, M.~M{\"u}ller, V.~Koltun, and
  D.~Scaramuzza, ``Champion-level drone racing using deep reinforcement
  learning,'' \emph{Nature}, vol. 620, no. 7976, pp. 982--987, 2023.

\bibitem{Hermansdorfer.5202020}
\BIBentryALTinterwordspacing
L.~Hermansdorfer, J.~Betz, and M.~Lienkamp, ``Benchmarking of a software stack
  for autonomous racing against a professional human race driver.'' [Online].
  Available: \url{http://arxiv.org/pdf/2005.10044v1}
\BIBentrySTDinterwordspacing

\bibitem{Kegelman.2018}
\BIBentryALTinterwordspacing
J.~C. Kegelman, ``Learning from professional race car drivers to make automated
  vehicles safer,'' Dissertation, {Stanford University}, Standford, California,
  2018. [Online]. Available: \url{https://purl.stanford.edu/jh569zw7186}
\BIBentrySTDinterwordspacing

\bibitem{Raji.2023}
A.~Raji, D.~Caporale, F.~Gatti, A.~Giove, M.~Verucchi, D.~Malatesta, N.~Musiu,
  A.~Toschi, S.~R. Popitanu, F.~Bagni, M.~Bosi, A.~Liniger, M.~Bertogna,
  D.~Morra, F.~Amerotti, L.~Bartoli, F.~Martello, and R.~Porta, ``er.autopilot
  1.0: The full autonomous stack for oval racing at high speeds.''

\bibitem{Betz.2023}
J.~Betz, T.~Betz, F.~Fent, M.~Geisslinger, A.~Heilmeier, L.~Hermansdorfer,
  T.~Hermann, S.~Huch, P.~Karle, M.~Lienkamp, B.~Lohmann, F.~Nobis,
  L.~{\"O}gretmen, M.~Rowold, F.~Sauerbeck, T.~Stahl, R.~Trauth, F.~Werner, and
  A.~Wischnewski, ``Tum autonomous motorsport: An autonomous racing software
  for the indy autonomous challenge,'' \emph{Journal of Field Robotics}, 2023.

\bibitem{Indy2023}
\BIBentryALTinterwordspacing
{Indy Autonomous Challenge}, ``Indy autonomous challenge sets autonomous speed
  records at monza ``temple of speed'','' 20.06.2023. [Online]. Available:
  \url{https://www.indyautonomouschallenge.com/}
\BIBentrySTDinterwordspacing

\bibitem{Cai.2021}
P.~Cai, H.~Wang, H.~Huang, Y.~Liu, and M.~Liu, ``Vision-based autonomous car
  racing using deep imitative reinforcement learning,'' \emph{IEEE Robotics and
  Automation Letters}, vol.~6, no.~4, pp. 7262--7269, 2021.

\bibitem{Fuchs.2021}
F.~Fuchs, Y.~Song, E.~Kaufmann, D.~Scaramuzza, and P.~Durr, ``Super-human
  performance in gran turismo sport using deep reinforcement learning,''
  \emph{IEEE Robotics and Automation Letters}, vol.~6, no.~3, pp. 4257--4264,
  2021.

\bibitem{Samak.10112021}
\BIBentryALTinterwordspacing
C.~V. Samak, T.~V. Samak, and S.~Kandhasamy, ``Autonomous racing using a hybrid
  imitation-reinforcement learning architecture.'' [Online]. Available:
  \url{http://arxiv.org/pdf/2110.05437v2}
\BIBentrySTDinterwordspacing

\bibitem{Samak.2021}
T.~V. Samak, C.~V. Samak, and S.~Kandhasamy, ``Robust behavioral cloning for
  autonomous vehicles using end-to-end imitation learning,'' \emph{SAE
  International Journal of Connected and Automated Vehicles}, vol.~4, no.~3,
  2021.

\bibitem{Cai.2020}
P.~Cai, X.~Mei, L.~Tai, Y.~Sun, and M.~Liu, ``High-speed autonomous drifting
  with deep reinforcement learning,'' \emph{IEEE Robotics and Automation
  Letters}, vol.~5, no.~2, pp. 1247--1254, 2020.

\bibitem{Betz.2022}
\BIBentryALTinterwordspacing
J.~Betz, H.~Zheng, A.~Liniger, U.~Rosolia, P.~Karle, M.~Behl, V.~Krovi, and
  R.~Mangharam, ``Autonomous vehicles on the edge: A survey on autonomous
  vehicle racing,'' \emph{IEEE Open Journal of Intelligent Transportation
  Systems}, vol.~3, pp. 458--488, 2022. [Online]. Available:
  \url{http://arxiv.org/pdf/2202.07008v1}
\BIBentrySTDinterwordspacing

\bibitem{Hao.11172022}
\BIBentryALTinterwordspacing
C.~Hao, C.~Tang, E.~Bergkvist, C.~Weaver, L.~Sun, W.~Zhan, and M.~Tomizuka,
  ``Outracing human racers with model-based planning and control for time-trial
  racing.'' [Online]. Available: \url{http://arxiv.org/pdf/2211.09378v2}
\BIBentrySTDinterwordspacing

\bibitem{Remonda.2021}
A.~Remonda, E.~Veas, and G.~Luzhnica, ``Comparing driving behavior of humans
  and autonomous driving in a professional racing simulator,'' \emph{PloS one},
  vol.~16, no.~2, p. e0245320, 2021.

\bibitem{Doubek.2021}
F.~Doubek, F.~Salzmann, and J.~de~Winter, ``What makes a good driver on public
  roads and race tracks? an interview study,'' \emph{Transportation Research
  Part F: Traffic Psychology and Behaviour}, vol.~80, pp. 399--423, 2021.

\bibitem{FrederikWerner.2024}
\BIBentryALTinterwordspacing
{Frederik Werner}, ``Github: Additional material,'' 2024. [Online]. Available:
  \url{https://github.com/TUM-AVS/AcceleratingAutonomy_ExpertInterviews_AdditionalMaterial}
\BIBentrySTDinterwordspacing

\bibitem{Mayring.2000}
P.~Mayring, ``Qualitative content analysis: Forum: Qualitative social research,
  vol 1, no 2 (2000): Qualitative methods in various disciplines i:
  Psychology,'' \emph{Forum: Qualitative Social Research}, 2000.

\bibitem{Kaiser.2014}
R.~Kaiser, \emph{Qualitative Experteninterviews}.\hskip 1em plus 0.5em minus
  0.4em\relax Wiesbaden: {Springer Fachmedien Wiesbaden}, 2014.

\bibitem{Milliken.1995}
W.~F. Milliken and D.~L. Milliken, \emph{Race car vehicle dynamics}, 14th~ed.,
  ser. SAE R.\hskip 1em plus 0.5em minus 0.4em\relax Warrendale, Pa.: {Society
  of Automotive Engineers}, 1995, vol. 146.

\bibitem{Acosta.2017}
M.~Acosta, S.~Kanarachos, and M.~Blundell, ``Road friction virtual sensing: A
  review of estimation techniques with emphasis on low excitation approaches,''
  \emph{Applied Sciences}, vol.~7, no.~12, p. 1230, 2017.

\bibitem{Khaleghian.2017}
S.~Khaleghian, A.~Emami, and S.~Taheri, ``A technical survey on tire-road
  friction estimation,'' \emph{Friction}, vol.~5, no.~2, pp. 123--146, 2017.

\bibitem{Wischnewski.2023b}
A.~Wischnewski, T.~Herrmann, F.~Werner, and B.~Lohmann, ``A tube-mpc approach
  to autonomous multi-vehicle racing on high-speed ovals,'' \emph{IEEE
  Transactions on Intelligent Vehicles}, vol.~8, no.~1, pp. 368--378, 2023.

\bibitem{Raji.2023b}
\BIBentryALTinterwordspacing
A.~Raji, N.~Musiu, A.~Toschi, F.~Prignoli, E.~Mascaro, P.~Musso, F.~Amerotti,
  A.~Liniger, S.~Sorrentino, and M.~Bertogna, ``A tricycle model to accurately
  control an autonomous racecar with locked differential.'' [Online].
  Available: \url{http://arxiv.org/pdf/2312.14808.pdf}
\BIBentrySTDinterwordspacing

\bibitem{Laurense.2017}
V.~A. Laurense, J.~Y. Goh, and J.~C. Gerdes, ``Path-tracking for autonomous
  vehicles at the limit of friction,'' in \emph{2017 American Control
  Conference (ACC)}.\hskip 1em plus 0.5em minus 0.4em\relax IEEE, 2017, pp.
  5586--5591.

\bibitem{Wischnewski.112019}
A.~Wischnewski, J.~Betz, and B.~Lohmann, ``A model-free algorithm to safely
  approach the handling limit of an autonomous racecar,'' in \emph{2019 IEEE
  International Conference on Connected Vehicles and Expo (ICCVE)}.\hskip 1em
  plus 0.5em minus 0.4em\relax IEEE, 11/4/2019 - 11/8/2019, pp. 1--6.

\bibitem{Kapania.2020}
N.~R. Kapania and J.~C. Gerdes, ``Learning at the racetrack: Data-driven
  methods to improve racing performance over multiple laps,'' \emph{IEEE
  Transactions on Vehicular Technology}, vol.~69, no.~8, pp. 8232--8242, 2020.

\end{thebibliography}

\end{document}